\documentclass[runningheads]{llncs}
\usepackage[T1]{fontenc}
\usepackage{graphicx,verbatim}
\usepackage{xcolor}
\usepackage{graphics}
\usepackage{epsfig} 
\usepackage{mathptmx} 
\usepackage{times}
\usepackage{amsmath,amssymb,amsfonts,bm}
\usepackage{algorithm,algorithmic}
\usepackage{flushend}
\usepackage{booktabs,threeparttable,multirow, caption}
\usepackage{color} 
\usepackage{cuted}
\usepackage{array}
\usepackage{tabu}                     
\usepackage{multirow}                
\usepackage{multicol}               
\usepackage{multirow}              
\usepackage{float}                   
\usepackage{makecell}               
\usepackage{booktabs} 
\usepackage{marvosym}

\begin{document}
% \newcommand{\qingyao}[1]{\textcolor{red}{\textbf{#1}}}

% \title{EndoMamba: Pre-training State Space Model for Real-Time Endoscopy Video Analysis}
\title{EndoMamba: An Efficient Foundation Model for Endoscopic Videos via Hierarchical Pre-training}

\titlerunning{EndoMamba: An Efficient Foundation Model for Endoscopic Videos}

%
\begin{comment}  %% Removed for anonymized MICCAI 2025 submission
\author{First Author\inst{1}\orcidID{0000-1111-2222-3333} \and
Second Author\inst{2,3}\orcidID{1111-2222-3333-4444} \and
Third Author\inst{3}\orcidID{2222--3333-4444-5555}}
%
\authorrunning{F. Author et al.}
\institute{Princeton University, Princeton NJ 08544, USA \and
Springer Heidelberg, Tiergartenstr. 17, 69121 Heidelberg, Germany
\email{lncs@springer.com}\\
\url{http://www.springer.com/gp/computer-science/lncs} \and
ABC Institute, Rupert-Karls-University Heidelberg, Heidelberg, Germany\\
\email{\{abc,lncs\}@uni-heidelberg.de}}

\end{comment}

\author{Qingyao Tian\inst{1,2}\and
Huai Liao\inst{3}\and
Xinyan Huang\inst{3}\and
Bingyu Yang\inst{1,2}\and
Dongdong Lei\inst{4}\and
Sebastien Ourselin\inst{5}\and
Hongbin Liu\inst{1,4,5}\textsuperscript{\Letter}
}

\authorrunning{Q. Tian et al.}

\institute{State Key Laboratory of Multimodal Artificial Intelligence Systems, Institute of Automation, Chinese Academy of Sciences, Beijing, China
\email{liuhongbin@ia.ac.cn}\and
School of Artificial Intelligence, University of Chinese Academy of Sciences, Beijing, China\and
The First Affiliated Hospital, Sun Yat-sen University, Guangzhou, China\and
Centre for Artificial Intelligence and Robotics, Chinese Academy of Sciences, HK, China\and
School of Engineering and Imaging Sciences, King’s College London, UK}

\maketitle              % typeset the header of the contribution
\begin{abstract}
%%%%%%%%%%%%%%%%%%%%%%%%%%%%%
Endoscopic video-based tasks, such as visual navigation and surgical phase recognition, play a crucial role in minimally invasive surgeries by providing real-time assistance. {While recent video foundation models have shown promise, their applications are hindered by \textit{(1) computational inefficiencies} and (2) suboptimal performance caused by \textit{limited data} for pre-training in endoscopy.} To address these issues, we present \textbf{EndoMamba}, a foundation model designed for real-time inference while learning generalized spatiotemporal representations. \textit{First}, to mitigate computational inefficiencies, we propose the EndoMamba backbone, optimized for real-time inference. Inspired by recent advancements in state space models, EndoMamba integrates Bidirectional Mamba blocks for spatial modeling within individual frames and vanilla Mamba blocks for past-to-present reasoning across the temporal domain. This design enables both strong spatiotemporal modeling and efficient inference in online video streams. \textit{Second}, we propose a self-supervised hierarchical pre-training diagram to enhance EndoMamba’s representation learning using endoscopic videos and incorporating general video domain knowledge. Specifically, our approach combines masked reconstruction with auxiliary supervision, leveraging low-level reconstruction to capture spatial-temporal structures and high-level alignment to transfer broader knowledge from a pretrained general-video domain foundation model. Extensive experiments on four downstream tasks—classification, segmentation, surgical phase recognition, and localization—demonstrate that EndoMamba outperforms existing foundation models and task-specific methods while maintaining real-time inference speed. The source code is available at \url{https://github.com/TianCuteQY/EndoMamba}. 

\keywords{Foundation model  \and Endoscopy video analysis \and State space model.}
% Authors must provide keywords and are not allowed to remove this Keyword section.

\end{abstract}
\section{Introduction}

Tasks based on endoscopic videos, such as navigation \cite{sganga2019autonomous} and surgical phase recognition \cite{liu2023skit}, have emerged as a key area of research in medical image analysis. These tasks contribute significantly to minimally invasive surgeries by providing real-time, automated assistance. The core challenge in endoscopic video analysis lies in the efficient estimation of spatiotemporal information \cite{li2024videomamba}. To improve performance, most research has focused on designing specialized modules to learn task specific domain knowledge \cite{liu2023skit,liu2025lovit,sganga2019autonomous}. 
However, task-specific methods often struggle with limited generalization and adaptability to new data, restricting their applicability in real clinical settings \cite{zhang2024generalist}.

Recently, foundation models have demonstrated remarkable performance in general video understanding \cite{tong2022videomae,wang2023videomaev2,li2023UMT}. These models have become the backbone of various video downstream applications, offering superior generalization and scalability compared to task-specific approaches \cite{wang2024internvideo2}. Recognizing their potential in endoscopy, EndoFM \cite{wang2023endofm} introduces a Transformer-based foundation model for endoscopic video analysis, achieving promising results on downstream tasks. However, two major challenges remain unaddressed: \textit{(1) computational inefficiencies} in online inference and \textit{(2) the limited quantity and deficient endoscopic data for effective pre-training} \cite{wang2023endofm}. The limitation in inference speed primarily stems from the backbone architecture. Existing video foundation models, such as Transformer-based models \cite{tong2022videomae,wang2023videomaev2} and bidirectional state space models (SSMs) \cite{li2024videomamba}, require recalculating all historical video frames repeatedly when new frames arrive during inference. This inefficiency significantly impacts real-time applications. Regarding data limitations, the main challenges are the relatively small dataset sizes and the lack of paired vision-language data in the endoscopic domain. These limitations restrict the potential for large-scale learning \cite{wang2023videomaev2} and contrastive learning \cite{radford2021clip,li2023UMT}, which are essential for improving model performance.
% Regarding the data limitation, a major issue is the absence of paired video-text endoscopic datasets. Without such datasets, endoscopic foundation models cannot leverage vision-text alignment for learning generalized representations, which is crucial for pretraining general vision models \cite{radford2021clip,li2023UMT} to improve performance.

% Closely related to our work, EndoFM \cite{wang2023endofm} introduces a Transformer-based foundation model for endoscopic video estimation, demonstrating promising results on downstream tasks. However, further efforts are needed to enable real-time applications and develop more effective pre-training strategies to learn generalizable representations.

In this work, we present EndoMamba, an efficient foundation model for endoscopy videos via hierarchical pre-training. To address the computational bottleneck, we propose the EndoMamba backbone. Motivated by recent advancements in SSMs \cite{gu2024mamba,dao2024mamba2} for long-term memory and fast inference, EndoMamba integrates Bidirectional Mamba (Bi-Mamba) for spatial modeling within individual frames and vanilla Mamba for causal reasoning across the temporal domain. This architecture enables both strong spatiotemporal reasoning and efficient inference in online video streams. To mitigate data limitations, we propose a hierarchical self-supervised pre-training diagram, enabling EndoMamba to capture spatiotemporal structures while leveraging broader knowledge from pretrained general-domain models. This diagram comprises two main components: low-level video reconstruction and high-level feature alignment. For the low-level reconstruction, inspired by VideoMAE \cite{tong2022videomae}, we adopt a masked autoencoder strategy for data-efficient pre-training, allowing EndoMamba to reconstruct video clips from heavily occluded visual cues. Additionally, we propose an auxiliary pre-training module for high-level feature alignment. This module leverages general-domain video foundation models trained on large-scale vision-text data to align features effectively. Extensive experiments are conducted on four downstream tasks. Results show that EndoMamba outperforms EndoFM with a +11.5\% increase in segmentation Dice score and a +21.3\% improvement in surgical phase recognition accuracy, while boosting inference speed from 9.2 frames per second (FPS) to 46.7 FPS with a memory length of 32 frames.

% Inspired by recent advancements in state space models (SSMs) for long-term memory and fast inference, we propose EndoMamba, a Mamba-based foundation model designed for real-time endoscopic video analysis with enhanced temporal reasoning. EndoMamba integrates Bi-Mamba blocks for spatial attention within individual frames and vanilla Mamba blocks for causal past-to-present reasoning across the temporal domain, enabling efficient inference in online video streams.
% To address training data limitations, we curate an expanded meta-dataset that doubles the size of EndoFM’s. Since vision-text data pairs are unavailable in endoscopy, we adopt a masked video modeling pre-training paradigm based on VideoMAE \cite{tong2022videomae}, where EndoMamba reconstructs video clips heavily masked in a tube-shaped pattern. Additionally, we introduce a teacher-student pre-training framework, leveraging general-domain video foundation models trained on large-scale vision-text data for feature alignment to enhance representation learning.
% Beyond classification and segmentation tasks, we extend experiments to surgical phase recognition and bronchoscopy branch-level localization, which require long-term memory. 
% Results show that EndoMamba outperforms existing endoscopic video foundation model, achieving a 15.3\% increase in segmentation Dice score and a 33.5\% improvement in surgical phase recognition accuracy, while boosting inference speed from 9.2 frames per second (FPS) to 46.7 FPS with a memory length of 32 frames.

\begin{figure*}[tbp]
\centerline{\includegraphics[width=\textwidth]{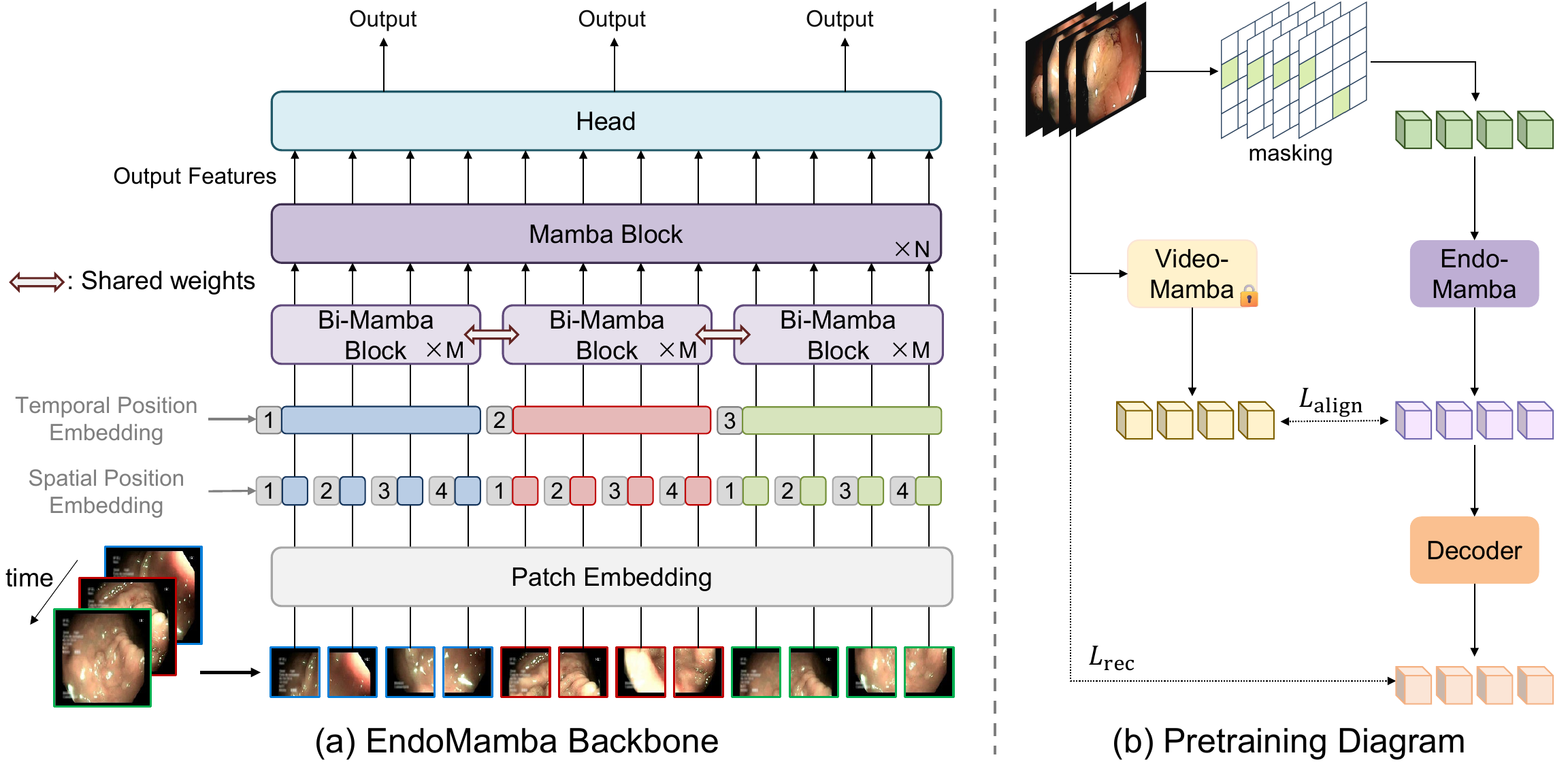}}
\caption{Proposed EndoMamba model for endoscopic video analysis. (a) EndoMamba backbone structure with efficient recurrent inference ability, and (b) hierarchical pre-training diagram for enhanced representation learning.}
\label{fig:diagram}
\end{figure*}

\section{Methods}

In this section, we first provide a brief overview of SSMs (\S 2.1). We then present a detailed description of the EndoMamba architecture (\S 2.2), which effectively captures generalized spatiotemporal representations while ensuring efficient inference. Finally, to mitigate the challenges associated with endoscopy data, we propose a pre-training strategy for EndoMamba that enhances its representation learning capability (\S 2.3).

% In this section, we first introduce \yuchen{the background of} SSMs. followed by the design of EndoMamba’s architecture for long-term memory and efficient inference.\yuchen{In this section, we first introduce the background of SSMs. Then, we provide a detailed description of the EndoMamba architecture, which effectively captures generalized spatiotemporal representations while ensuring efficient inference.
% Finally, to address limitations in endoscopy data, we introduce the pre-training strategy on EndoMamba, which will enhance the model xxx } Finally, we present our pre-training framework, which incorporates advanced self-supervision and more challenging pre-training tasks to enhance representation learning.
% \yuchen{The goal of EndoMamba is to xxx. In this section, we first introduce the preliminaries of SSMs. Then, xxx}

\subsection{Preliminaries}
Deep State Space Models (SSMs), such as Mamba \cite{gu2024mamba}, have recently achieved success in natural language processing due to their ability to model long-range dependencies while maintaining computational efficiency. It is grounded by classical SSMs which map a one-dimensional function or sequence \( x(t) \in \mathbb{R} \) to an output \( y(t) \in \mathbb{R} \) through a hidden state \( h(t) \in \mathbb{R}^{{N}} \). The system dynamics are modeled by the evolution matrix \( \bm{A} \in \mathbb{R}^{{N} \times N} \), with projection parameters \( \bm{B} \in \mathbb{R}^{{N} \times 1} \) and \( \bm{C} \in \mathbb{R}^{1 \times N} \). The continuous system operates as follows:
\begin{equation}
    h^{\prime}(t) = \bm{A} h(t) + \bm{B} x(t), 
    y(t) = \bm{C} h(t).
\end{equation}

For efficient training with discrete data, SSMs discretizes the continuous system into a discrete form using the zero-order hold (ZOH) method, defined as:

% Mamba discretizes the continuous system by introducing a timescale parameter \( {\Delta} \) to convert the continuous parameters \( \bm{A}, \bm{B} \) into their discrete counterparts \( \bar{\bm{A}}, \bar{\bm{B}} \). The zero-order hold (ZOH) method is typically used for this transformation, defined as:
% \begin{equation}
% \bar{\bm{A}} = \exp({\Delta} \bm{A}), 
% \bar{\bm{B}} = ({\Delta} \bm{A})^{-1} \left( \exp({\Delta} \bm{A}) - \bm{I} \right) \cdot {\Delta} \bm{B}.
% \end{equation}

% \noindent Once discretized, the system with a step size \( \boldsymbol{\Delta} \) can be rewritten as:
\begin{equation}
h_t = \bar{\bm{A}} h_{t-1} + \bar{\bm{B}} x_t, 
y_t = \bm{C} h_t,
\label{eq:recurrent}
\end{equation}

\noindent where the discrete counterparts $\bar{\bm{A}}$, $\bar{\bm{B}}$ are defined as:

\begin{equation}
\bar{\bm{A}} = \exp({\Delta} \bm{A}), 
\bar{\bm{B}} = ({\Delta} \bm{A})^{-1} \left( \exp({\Delta} \bm{A}) - \bm{I} \right) \cdot {\Delta} \bm{B}.
\end{equation}

% \noindent The output can also be computed through a global convolution, represented as:
% \begin{equation}
% \bar{\bm{K}} = \left( \bm{C} \bar{\bm{B}}, \bm{C} \bar{\bm{A B}}, \dots, \bm{C} \bm{A}^{M-1} \bar{\bm{B}} \right), 
% \bm{y} = \bm{x} * \bar{\bm{K}},
% \label{eq:parellel}
% \end{equation}
% \noindent where \( M \) is the length of the input sequence \( \bm{x} \), and \( \bar{\bm{K}} \in \mathbb{R}^M \) is a structured convolution kernel.

Based on the SSMs, Mamba further proposes to selectively propagate or forget information along the sequence based on the current token, enabling efficient linear scalability in sequence length and strong long-range dependency modeling capability.

However, since vanilla Mamba processes input tokens in a past-to-present manner, it is not well-suited for vision tasks. To address this limitation, VisionMamba \cite{zhu2024visionmamba} uses the Bi-Mamba block, which applies bidirectional scanning to process the sequences from both forward and backward directions for image analysis. 

% This block is also effective for video processing tasks as it captures both spatial and temporal relationships \cite{li2024videomamba}.

% This block is also incorporated into video processing networks to capture both spatial and temporal relationships \cite{li2024videomamba}.

\subsection{EndoMamba Architecture}
Motivated by recent advancements in SSMs, we propose EndoMamba, a Mamba-based backbone designed to enhance computational efficiency in video foundation models. An overview of the proposed EndoMamba is shown in Fig. \ref{fig:diagram} (a). 
% \yuchen{overflow here, EndoMamba contains xx components. First, EndoMamba takes input xxxx, Finally, output}
As mamba blocks take input as 1-D sequence, we first transform the input videos $\bm{X} \in \mathbb{R}^{B \times 3 \times T \times H \times W}$ into flattened spatiotemporal patches $\bm{X}_p \in \mathbb{R}^{B \times N \times C}$ where $N=T \times h \times w\left(h=\frac{H}{k}\right.$, and $\left.w=\frac{W}{k}\right)$. Here, $k$ denotes the kernel size, $B$ is the batch size, $T$ denotes the number of frames, and $H \times W$ represents frame spatial resolution.
% \yuchen{The definition of $h, ...$}

To retain spatiotemporal position information, we add a learnable spatial position embedding $\bm{p}_s$, and a fixed sinusoid temporal embedding $\bm{p}_t$ to capture temporal position information while allowing flexible input sequence length. The embedded input is then:

\begin{equation}
\bm{X}_{\mathrm{embed}} = \bm{X}_p + \bm{p}_s + \bm{p}_t.
\end{equation}

To perform spatial-temporal scanning while retaining recurrent inference, EndoMamba process $\bm{X}_{\mathrm{embed}}$ by performing bidirectional scanning within each image patch and causal scanning along the time axis. Specifically, the embedded input $ \bm{X}_{\mathrm{embed}} $ are first rearranged into $ \bm{X}_r \in \mathbb{R}^{J \times P \times C} $, where $ J = B \times T $ and $ P = h \times w $. Then, $\bm{X}_r$ is passed through $ M $ stacked Bi-Mamba blocks for bidirectional scanning within each image frame in parallel.
After processing, the tokens are rearranged back to $ \bm{X}_b \in \mathbb{R}^{B \times N \times C} $ and passed through $ n $ stacked Mamba blocks, which perform causal scanning along the temporal axis. Finally, the processed tokens are fed into a multi-layer perceptron (MLP) head to produce the prediction. For frame-level tasks, the MLP head outputs $ \bm{Y} \in \mathbb{R}^{B \times T} $. 

For training, EndoMamba takes video clips as input and simultaneously outputs predictions for each frame, using information from the current frame as well as its past frames. For real-time applications, EndoMamba processes each frame of a live video stream individually, passing the past states as memory. The model then makes a final prediction for the current frame based on both the frame itself and its historical context.

\subsection{Pretraining Diagram}
To tackle the limitation of endoscopy data, we propose a hierarchical pre-training diagram for EndoMamba, as illustrated in Fig. \ref{fig:diagram}(b). This pre-training combines low-level video reconstruction with high-level feature alignment to enhance representation learning. For the low-level reconstruction, we randomly mask a large portion of the input video, prompting EndoMamba to infer the missing regions. This process inherently encourages EndoMamba to capture contextual dependencies by leveraging spatiotemporal correlations \cite{tong2022videomae}. Formally, the reconstruction loss is defined as follows:

\begin{equation}
    L_\mathrm{rec} = \frac{1}{|\Omega|} \sum_{p \in \Omega} \|\bm{X}(p) - \hat{\bm{X}}(p)\|^2,
\end{equation}

\noindent where $\Omega$ is the set of masked video regions, $p$ is the masked token index, and $\hat{\bm{X}}$ is the reconstructed video clip by EndoMamba.

To further enhance representation learning, we propose to align EndoMamba’s features with VideoMamba, a general-domain video model pretrained on large-scale video data as teacher model. The aligning is enforced by cosine similarity loss defined as:

\begin{equation}
    L_\mathrm{align}=1-\frac{1}{\bar{\Omega}} \sum_{q \in \bar{\Omega}}\cos \left(\bm{X_f}(q), \bm{X_t}(q)\right),
\end{equation}

\begin{equation}
    \cos\left(\bm{X_f}(q), \bm{X_t}(q)\right) = \frac{\bm{X_f}(q) \cdot \bm{X_t}(q)}{\|\bm{X_f}(q)\| \|\bm{X_t}(q)\|},
\end{equation}

\noindent where $\bm{X_f}$ and $\bm{X_t}$ are the output features of EndoMamba and the teacher model respectively; $\bar{\Omega}$ is the set of unmasked video regions and $q$ is the unmasked token index. 

% The cosine similarity is only calculated on the output layer of the backbone as EndoMamba utilizes its unique network architecture.

\begin{table}[tbp]
  \centering
  \caption{Overview of the pre-training dataset. Our full pre-training dataset MIX12 consists of 74,828 video clips, comprising a total of 11,303,853 frames.}
  \setlength{\tabcolsep}{2.8pt} 
    \begin{tabular}{c|ccccccc|ccccc}
    \hline
    & \multicolumn{7}{c|}{MIX7}                             & \multicolumn{5}{c}{Additional to MIX12} \\
    \hline
    Source & \thead{CS \\ \cite{mesejo2016computer}} & \thead{S\&S \\ \cite{misawa2021development,ji2022video}} & \thead{LP \\ \cite{ma2021ldpolypvideo}} & \thead{HK \\ \cite{borgli2020hyperkvasir}} & \thead{KC \\ \cite{smedsrud2021kvasir}} & \thead{CT \\ \cite{nwoye2022rendezvous}} & \thead{EFM \\ \cite{wang2023endofm}} & \thead{G \\ \cite{Andreas2020GLENDA}} & \thead{hS \\ \cite{yoon2021hsdb}} & \thead{EM \\ \cite{azagra2023endomapper}} & \thead{R \\ \cite{ross2021comparative}} & PB \\
    \hline
     Clips  & 210 & 1.0k & 237 & 5.7k & 1.0k & 580 & 24.1k & 104 & 236 & 1.7k & 39.8k & 179 \\
    Frames & 36.5k & 159.4k & 40.2k & 876k & 159k & 90.4k & 3.66M & 13.4k & 35.6k & 242k & 5.96M & 26.8k \\
    \hline
    \end{tabular}%
  \label{tab:datasets}%
\end{table}%

\begin{figure*}[tbp]
\centerline{\includegraphics[width=\textwidth]{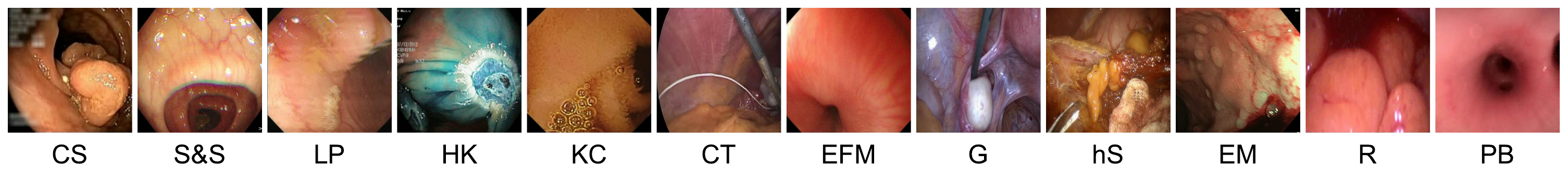}}
\caption{Example frames of the pre-training data in MIX12.}
\label{fig:dataset}
\end{figure*}

This feature alignment enables EndoMamba to inherit knowledge from a broader domain while introducing additional challenges in the pretraining process. Unlike the teacher model, EndoMamba is trained on significantly sparser inputs. Spatially, it operates on heavily masked video clips, while VideoMamba utilizes the full input. Temporally, VideoMamba leverages Bi-Mamba, enabling bidirectional scans across the temporal axis, whereas EndoMamba follows a causal scanning. Aligning these features thus necessitates a dual-axis completion of a robust general-domain representation.

Finally, the pre-training of EndoMamba uses the following loss function:

\begin{equation}
    L=L_\mathrm{rec} + \alpha L_\mathrm{align},
\end{equation}

\noindent where $\alpha$ is the weight for feature alignment loss. Our pre-training diagram allows the model to capture contextual dependencies and inherit knowledge from a broader domain for enhanced generalization ability.

\section{Experiments}

\subsection{Implementation Details}
To increase data diversity and volume, EndoMamba is pretrained on MIX12, a combination of 12 endoscopic datasets. This includes MIX7 (shared with EndoFM), four curated from public datasets, and one from our porcine bronchoscopy dataset (PB), as detailed in Table \ref{tab:datasets} and Fig. \ref{fig:dataset}. This increases the pre-training data from approximately 5M to 11M frames when moving from MIX7 to MIX12. The model is trained using the AdamW optimizer with a base learning rate of 1e-8, a cosine learning rate schedule for 500 epochs, and a batch size of 48, with the first five epochs dedicated to linear warm-up. For EndoMamba architecture, we set M=12, N=12. Frames in each video clips are resized to spatial size of 224×224 and frame number of 16 as pre-training input. For feature alignment, the pretrained VideoMamba-S serves as the teacher model, with the feature alignment loss weight set to $\alpha = 0.25$ based on preliminary experiments.

To evaluate the generalization ability of EndoMamba, we transfer it to a range of downstream tasks, including 1) \textit{PolypDiag for classification} \cite{tian2022contrastive} and 2) \textit{CVC-12K for polyp segmentation} \cite{bernal2015wm}, using the same fine-tuning settings as EndoFM. Additionally, we extend its application to two other tasks: 3) \textit{surgical phase recognition on the Autolaparo dataset} \cite{wang2022autolaparo} and 4) \textit{branch-level localization on patient bronchoscopy dataset}. Compared to classification and segmentation, surgical phase recognition and localization emphasis long-term memory and real-time inference. For the latter two tasks, we sample 32-frame input clips of spatial size 224×224 for fine-tuning, with data splits following existing methods \cite{liu2023skit,tian2024endoomni}. The number of training epochs is set to 50 for surgical phase recognition and 20 for localization. No task-specific modules are used; instead, only an MLP head is added to the EndoMamba backbone for final output.

\subsection{Comparison with State of the Art}
We compare EndoMamba with recent state-of-the-art (SOTA) methods on the four downstream tasks. \textbf{1) Classification:} Compared to existing pre-training methods and foundation models \cite{ding2022fame,park2022ProViCo,qian2022vcl,pan2022stadapter,wang2023endofm,wang2023videomaev2,li2024videomamba} , EndoMamba ranks first (Table \ref{tab:result}), achieving a +4.3\% improvement in F1 score over the best-performing SOTA method. \textbf{2) Segmentation:} EndoMamba maintains its leading performance in segmentation (Table \ref{tab:result}) among compared pre-training methods and foundation models \cite{ding2022fame,park2022ProViCo,qian2022vcl,pan2022stadapter,wang2023endofm,wang2023videomaev2,li2024videomamba}, achieving a +6.9\% increase in Dice score over the best SOTA model. \textbf{3) Surgical Phase Recognition:} As shown in Table \ref{tab:surgicalphase}, EndoMamba outperforms all evaluated foundation models \cite{wang2023videomaev2,li2024videomamba,wang2023endofm}, achieving a +2.7\% increase in video-level accuracy over VideoMamba. Additionally, it performs competitively with the top task-specific methods \cite{gao2021transsvnet,girdhar2021avt,liu2025lovit,liu2023skit}, such as SKiT, which employs a multi-stage framework, whereas EndoMamba achieves comparable results using only its backbone and an MLP head for end-to-end estimation. \textbf{4) Localization:} EndoMamba outperforms all evaluated foundation \cite{tian2024endoomni,wang2023videomaev2,li2024videomamba,wang2023endofm} and task-specific models \cite{sganga2019autonomous,tian2024bronchotrack} by a significant margin (Table \ref{tab:localization}). Bronchoscopy interventions often face severe visual occlusions, such as liquids and bubbles, making accurate localization reliant on strong spatiotemporal reasoning. EndoMamba effectively recovers positions after occlusions, as shown by reduced video-level accuracy variance, indicating more stable performance across cases. These results demonstrate its ability to capture transferable representations robust enough to excel in complex downstream tasks without task-specific architectural modifications.

\begin{table}[t]
    \centering
    \begin{minipage}{0.52\textwidth}
        \centering
        \caption{Results of classification on PolypDiag \cite{tian2022contrastive} and segmentation on CVC-12k \cite{bernal2015wm}. \textbf{Best} and \underline{second best} results are highlighted.}
    \begin{tabular}{ccc}
    \hline
    \multirow{2}[4]{*}{Methods} & PolypDiag & CVC-12K \\
\cmidrule{2-3}          & F1 (\%) ↑ & Dice (\%) ↑\\
    \hline
    FAME \cite{ding2022fame} & 85.4±0.8 & 67.2±1.3 \\
    ProViCo \cite{park2022ProViCo} & 86.9±0.5 & 69.0±1.5 \\
    VCL \cite{qian2022vcl} & 87.6±0.6 & 69.1±1.2 \\
    ST-Adapter \cite{pan2022stadapter} & 84.8±0.7 & 64.3±1.9 \\
    EndoFM \cite{wang2023endofm} & \underline{90.7±0.4} & 73.9±1.2 \\
    VideoMAEv2 \cite{wang2023videomaev2} & 87.5±1.6 & 72.1±0.9 \\
    VideoMamba \cite{li2024videomamba} & 75.6±1.9 & \underline{78.5±1.0} \\
    Ours & \textbf{95.0±1.3} & \textbf{85.4±0.2} \\
    \hline
    \end{tabular}%
    \label{tab:result}
    \end{minipage}\hfill
    \hspace{0.35cm}%
    \begin{minipage}{0.45\textwidth}
        \centering
        \includegraphics[width=\linewidth]{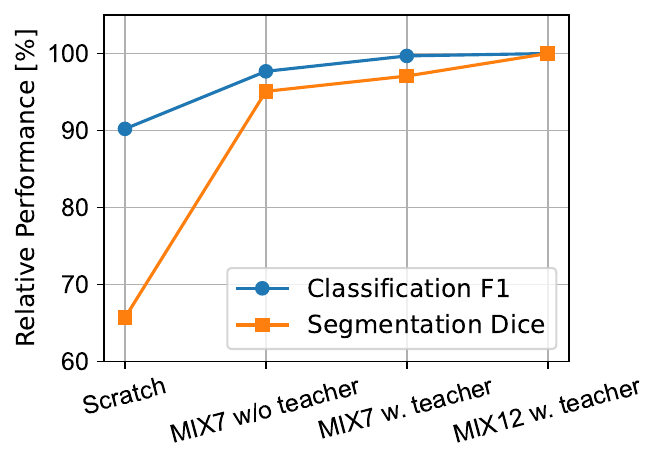}
        \captionof{figure}{Relative performance on classification F1 and segmentation Dice score, with gradual performance gains from pre-training data scaling and the addition of a teacher model.}
        \label{fig:ablation_modules}
    \end{minipage}
\end{table}

\begin{table}[tp]
    \centering
    \begin{minipage}{0.49\textwidth}
        \centering
        \caption{Surgical phase recognition results on the Autolaparo \cite{wang2022autolaparo} dataset, comparing task-specific methods (top) and foundation models (bottom).}
        \setlength{\tabcolsep}{4pt} 
    \begin{tabular}{ccc}
    \hline
    Method & Accuracy ↑ & Jaccard ↑ \\
    \midrule
    Trans-SVNet \cite{gao2021transsvnet} & 78.3  & 50.7 \\
    AVT \cite{girdhar2021avt} & 77.8  & 50.7 \\
    LoViT \cite{liu2025lovit} & 81.4 ± 7.6 & 56 \\
    SKiT \cite{liu2023skit} & \underline{82.9 ± 6.8} & \textbf{59.9} \\
    \midrule
    VideoMAEv2 \cite{wang2023videomaev2} & 77.0 ± 5.7 & 53.1  \\
    VideoMamba \cite{li2024videomamba} & 80.3 ± 8.2 & 54.0  \\
    EndoFM \cite{wang2023endofm} & 62.0 ± 11.9 & 37.4  \\
    Ours  & \textbf{83.3 ± 8.7} & \underline{57.6}  \\
    \bottomrule
    \end{tabular}%
  \label{tab:surgicalphase}%
    
    \end{minipage}\hfill
    \hspace{0.5cm}%
    \begin{minipage}{0.45\textwidth}
        \caption{Localization results on patient bronchoscopy dataset, comparing task-specific methods (top) and foundation models (bottom).}
        \setlength{\tabcolsep}{4pt} 
    \begin{tabular}{ccc}
    \hline
    Methods & Accuracy ↑ & F1 ↑ \\
    \midrule
    AirwayNet \cite{sganga2019autonomous} & 38.7 ± 15.7 & 53.1  \\
    BronchoTrack \cite{tian2024bronchotrack} & 57.0 ± 24.6 & 61.3  \\
    \midrule
    EndoOmni \cite{tian2024endoomni} & 78.6 ±16.2 & \underline{69.2}  \\
    VideoMAEv2 \cite{wang2023videomaev2} & \underline{78.9 ± 9.7} & \underline{69.2}  \\
    VideoMamba \cite{li2024videomamba} & 74.1 ± 8.3 & 67.4  \\
    EndoFM \cite{wang2023endofm} & 62.2 ± 19.1 & 56.7  \\
    Ours  & \textbf{83.0 ± 5.5} & \textbf{73.5} \\
    \bottomrule
    \end{tabular}%
    \label{tab:localization}
    \end{minipage}
\end{table}

\begin{table}[tp]
  \centering
  \caption{Inference speed comparison of EndoMamba with existing video foundation models. Results are reported in frames per second (FPS) with network memory lengths set to T=32, 64, 128. Inference complexity \cite{dao2024mamba2} is measured by the number of image tokens ($P$), network memory lengths ($T$) and hidden state dimension ($m$).}
  \setlength{\tabcolsep}{5pt}
\begin{tabular}{ccccccc}
    \toprule
    \multirow{2}{*}[-0.3ex]{\centering Method} & 
    \multirow{2}{*}[-0.3ex]{\centering Backbone} & 
    \multirow{2}{*}[-0.3ex]{\centering Param Num.} & 
    \multirow{2}{*}[1ex]{\centering \thead{Inference \\ Complexity}} & 
    \multicolumn{3}{c}{FPS↑} \\
\cline{5-7}          &       &       &       & T=32  & T=64  & T=128 \\
    \midrule
    EndoFM \cite{wang2023endofm} & video ViT & 121.26M &  $O(P^2T^2m)$   & 9.2   & 4.8   & 2.4  \\
    VideoMAEv2 \cite{wang2023videomaev2} & video ViT & 22.26M & $O(P^2T^2m)$      & 27.8  & 9.7   & 2.7  \\
    VideoMamba \cite{li2024videomamba} & VideoMamba & 25.42M &  $O(PTm^2)$     & 21.7  & 11.8  & 6.1  \\
    Ours  & EndoMamba & 24.46M &    $O(Pm^2)$   & \textbf{46.7 } & \textbf{47.1 } & \textbf{46.6 } \\
    \bottomrule
    \end{tabular}%
% }
   \label{tab:speed}%
\end{table}%

\subsection{Ablation Studies}
We explore various aspect of EndoMamba, regarding pre-training settings and real-time performance. \textbf{1) Hierachical Pre-training:} We compare training from scratch with hierarchical pre-training, by evaluating downstream performance on classification and segmentation. As seen in Fig. \ref{fig:ablation_modules}, "\textit{MIX12 w. teacher}", representing our full pre-training strategy, significantly outperforms "\textit{Scratch}", which denotes random initialization. The improvement highlights the effectiveness of our pre-training, which includes low-level video reconstruction for learning spatiotemporal structure and high-level alignment for enhanced performance.
\textbf{2) Teacher Guidence:} As seen in Fig. \ref{fig:ablation_modules}, "\textit{MIX7 w. teacher}", which represents pre-training on MIX7 dataset with the teacher model guidance, brings an improvement of around 2\% on the evaluated two tasks, comparing with "\textit{MIX7 w/o teacher}", which uses the same dataset without teacher guidance. This improvement stems from high-level feature alignment, which transfers broader knowledge from a pretrained general-domain foundation model, thereby yielding more generalizable representations.
\textbf{3) Dataset Scaling:} To explore the improvement brought by pre-training data scaling, we compare the performance of pre-training with MIX7, which aligns with existing endoscopy video foundation model EndoFM, and pre-training using our expanded MIX12. Results are shown in Fig. \ref{fig:ablation_modules} "\textit{MIX7 w. teacher}" and "\textit{MIX12 w. teacher}" respectively. Dataset scaling up contribute to 3\% improvement on the more challenging segmentation task, and a moderate improvement on classification. \textbf{4) Speed Analysis:} 
To evaluate the real-time applicability of EndoMamba, we compare its speed with existing foundation models, including the transformer-based EndoFM and VideoMAE v2, as well as the Mamba-based VideoMamba. Results for inference speed, computational complexity, and parameter counts of all models are presented in Table \ref{tab:speed}. For a fair comparison, inference speed is measured in frames per second (FPS) on an NVIDIA A800 GPU with memory lengths of 32, 64, and 128 frames. Results show that EndoMamba achieves the highest inference speed among all tested models with the evaluated memory length, significantly outperforming both transformer-based and Mamba-based architectures. Notably, its inference efficiency remains stable as memory size increases, benefiting from its ability to propagate past memory without redundant computation. In contrast, transformer-based models suffer from quadratic-complexity attention mechanisms, while Bi-Mamba blocks require recomputation, both of which hinder efficiency. By maintaining temporal continuity with minimal computational overhead, EndoMamba is particularly well-suited for real-time endoscopic applications.

\section{Conclusion}
In this work, we present EndoMamba, a foundation model designed for real-time endoscopic video analysis. The proposed EndoMamba backbone employs spatial bidirectional scanning and temporal causal scanning, enabling strong spatiotemporal modeling and efficient inference. To address the data limitation, we propose a hierarchical pre-training diagram that combines low-level video reconstruction for spatiotemporal representation learning, and high-level alignment with a pretrained general-domain foundation model, leveraging its broader visual knowledge to enhance representation learning. Experimental results on 4 diverse downstream tasks demonstrate that EndoMamba outperforms existing foundation models and SOTA task-specific methods while maintaining real-time inference efficiency for practical deployment.
\bibliographystyle{splncs04}
\bibliography{references}
\end{document}